\documentclass[11pt,letterpaper]{article}
\usepackage{arabtex}
\usepackage{utf8}

\usepackage{naaclhlt2016}
\usepackage{times}
\usepackage{latexsym}
\usepackage{graphicx}
\usepackage{enumitem}

\naaclfinalcopy 
\def\naaclpaperid{***} 

\title{Capturing Reliable Fine-Grained Sentiment Associations\\ by Crowdsourcing and Best--Worst Scaling}

\author{Svetlana Kiritchenko \and Saif M. Mohammad\\
	    National Research Council Canada\\
	    {\tt \small \{svetlana.kiritchenko,saif.mohammad\}@nrc-cnrc.gc.ca}
}

\date{}

\begin{document}

\maketitle

\newcommand{\measure}[2]{$\rm \it #1_{{\rm \it #2}}$}
\newcommand{\score}[1]{$\rm \it score({\rm \it #1})$}

\begin{abstract}
Access to word--sentiment associations is useful for many applications, including sentiment analysis, stance detection, and linguistic analysis.  
However, manually assigning fine-grained sentiment association scores to words has many challenges with respect to keeping annotations consistent. 
We apply the annotation technique of Best--Worst Scaling 
to obtain real-valued sentiment association scores for words and phrases in three different domains: general English, English Twitter, and Arabic Twitter.  
We show that on all three domains the ranking of words by sentiment remains remarkably consistent even when the annotation process is repeated with a different set of annotators.
We also, for the first time, determine the minimum difference in sentiment association that is perceptible to native speakers of a language.
\end{abstract}

\section{Introduction}
\setitemize[0]{leftmargin=*}
\setenumerate[0]{leftmargin=*}
Word--sentiment associations, commonly captured in sentiment lexicons, are useful in automatic sentiment prediction \cite{Semeval2014task4,rosenthal-EtAl:2014:SemEval}, 
stance detection \cite{stance-lrec,MohammadSK16}, 
literary analysis \cite{hartner2013lingering,kleres2011emotions,Mohammad2012},
detecting personality traits \cite{grijalva2014gender,COIN:COIN12024}, 
and other applications. 
Manually created sentiment lexicons are especially useful because they tend to be more accurate than automatically generated lexicons; 
they can be used to automatically generate large-scale lexicons 
\cite{tang2014building,Esuli06}; 
they can be used to evaluate different methods of automatically creating sentiment lexicons; 
and they can be used for linguistic analyses such as examining how sentiment is composed in phrases and sentences. 

The sentiment of a phrase can differ significantly from the sentiment of its constituent words. 
Sentiment composition is the determining of sentiment of a multi-word linguistic unit, such as a phrase or a sentence, from its constituents. 
Lexicons that include sentiment associations for phrases as well as for their constituent words are useful in studying sentiment composition. 
We refer to them as {\it sentiment composition lexicons (SCLs)}. 
We created SCLs for three domains, and all three were used in recent SemEval shared tasks. We refer to the lexicon created for the English Twitter domain as the {\it SemEval-2015 English Twitter Sentiment Lexicon};
for the general English domain as the {\it SemEval-2016 General English Sentiment Modifiers Lexicon};
and for the Arabic Twitter domain as the {\it SemEval-2016 Arabic Twitter Sentiment Lexicon}. 
Note that the English Twitter lexicon was first described in \cite{Kiritchenko2014}, whereas the other two are novel contributions presented in this paper.

Most existing manually created sentiment lexicons tend to provide only lists of positive and negative words with very coarse levels of sentiment
\cite{Stone66,Hu04,Wilson05,MohammadT13}.  
The coarse-grained distinctions may be less useful in downstream applications than having access to fine-grained (real-valued) sentiment association scores. 
Only a small number of manual lexicons capture sentiment associations at a fine-grained level  \cite{bradley1999affective,warriner2013norms}. 
This is not surprising because obtaining real-valued sentiment annotations has several challenges.  
Respondents are faced with a higher cognitive load when asked for real-valued sentiment scores for terms as opposed to simply classifying terms as either positive or negative.
Besides, it is difficult for an annotator to remain consistent with his/her annotations.
Further, the same sentiment association may map to different sentiment scores in the minds of different annotators; for example, one annotator may assign a score of 0.6 and another 0.8 for the same degree of positive association.
One could overcome these problems by providing annotators with pairs of terms and asking which is more positive (a comparative approach), however that requires a much larger set of annotations (order $N^2$, where N is the number of terms to be annotated).
Best--Worst Scaling (BWS) is an annotation technique, commonly used in marketing research
\cite{Louviere_1990}, that exploits the comparative approach to annotation while keeping the number
of required annotations small.  

In this work, we investigate the applicability and reliability of the Best--Worst Scaling annotation technique in capturing word--sentiment associations via crowdsourcing. 
Our main contributions are as follows:
\vspace*{-2mm}
\begin{enumerate} 
\item We create fine-grained sentiment composition lexicons for Arabic Twitter and general English (in addition to our earlier work on English Twitter) using Best--Worst Scaling. 
The lexicons include entries for single words as well as multi-word phrases. 
The sentiment scores are real values between -1 (most negative) and +1 (most positive).
\vspace*{-2mm}
\item We show that the annotations on all three domains are reliable; 
re-doing the annotation  with different sets of annotators produces a very similar order of terms---an average Spearman rank correlation of 0.98. 
	Furthermore, we show that reliable rankings can be obtained with just two or three annotations per BWS question. 
(\newcite{warriner2013norms} and \newcite{grahamaccurate} have shown that conventional rating-scale methods require a much higher number of responses (15 to 20)).
\vspace*{-2mm}
\item We examine the relationship \mbox{between `differen-} ce in the sentiment scores between two terms' and `agreement amongst annotators' when asked which term is more positive.
We show that agreement grows rapidly and reaches 90\% when the difference in sentiment scores is about 0.4 (20\% of interval between -1 and 1).
\vspace*{-2mm}
\item  We calculate the minimum difference \mbox{in sentime-} nt scores of two terms that is perceptible to native speakers of a language. 
For sentiment scores between -1 (most negative) and 1 (most positive), we show that the perceptible difference is about 0.08 for English and Arabic speakers.
Knowing the least perceptible difference helps researchers better understand sentiment composition. For example, consider the task of determining
whether an adjective significantly impacts the sentiment of the noun it qualifies. This can be accomplished by determining whether the difference in sentiment scores between the combined phrase and
the constituent noun alone is greater than the least perceptible difference.
\end{enumerate} 
\vspace*{-2mm}

\noindent 
The data and code created as part of this project (the lexicons, the annotation questionnaire, and the code to generate BWS questions) are made available.\footnote{www.saifmohammad.com/WebPages/BestWorst.html}

\section{Capturing Fine-Grained Sentiment Associations By Manual Annotation}
\label{lexicon-description}

We now describe how we created three lexicons, through manual annotation, that each provide real-valued sentiment association scores.

\subsection{Best--Worst Scaling Method of Annotation} 
\label{maxdiff}

Best--Worst Scaling (BWS), also sometimes referred to as Maximum Difference Scaling (MaxDiff), is an annotation scheme that exploits the comparative approach to annotation  \cite{Louviere_1990,Cohen_2003,Louviere2015}. 
Annotators are given four items (4-tuple) and asked which item is the Best (highest in terms of the property of interest) and which is the Worst (least in terms of the property of interest). 
These annotations can then be easily converted into real-valued scores
of association between the items and the property, which eventually allows for creating a ranked list of items as per their association with the property of interest. 

Given $n$ terms to be annotated, the first step is to randomly sample this set (with replacement) to obtain sets of four terms each, {\it 4-tuples}, that satisfy the following criteria:  
\setenumerate[0]{leftmargin=8mm}
\begin{enumerate}
\vspace*{-3mm}
\item no two 4-tuples have the same four terms; 
\vspace*{-3mm}
\item no two terms within a 4-tuple are identical; 
\vspace*{-3mm}
\item each term in the term list appears approximately in the same number of 4-tuples; 
\vspace*{-3mm}
\item each pair of terms appears approximately in the same number of 4-tuples. 
\end{enumerate}
\vspace*{-3mm}
In practice, around $1.5\times n$ to $2\times n$ BWS questions, where $n$ is the number of items, are sufficient to obtain reliable scores. 
We annotated terms for the three lexicons separately, and generated $2 \times n$ 4-tuples for each set. 

Next, the sets of 4-tuples were annotated through a crowdsourcing platform, CrowdFlower. 
The annotators were presented with four terms at a time, and asked which term is the most positive (or least negative) and which is the most negative (or least positive). 
Below is an example annotation question.\footnote{The full sets of instructions for both English and Arabic datasets are available at:\\ http://www.saifmohammad.com/WebPages/BestWorst.html} (The Arabic data was annotated through a similar questionnaire in Arabic.)
\vspace*{-1mm}
{\small
\noindent\makebox[\linewidth]{\rule{0.49\textwidth}{0.4pt}}\\
Focus terms:\\
1. th*nks \hspace{2mm} 2. doesn't work \hspace{2mm} 3. w00t \hspace{2mm} 4. \#theworst

\vspace*{1mm}
 \noindent Q1: Identify the term that is associated with the most amount of positive sentiment (or least amount of negative sentiment) -- \textbf{the most positive term}:\\
1. th*nks \hspace{2mm} 2. doesn't work \hspace{2mm} 3. w00t \hspace{2mm} 4. \#theworst

\vspace*{1mm}
 \noindent Q2: Identify the term that is associated with the most amount of negative sentiment (or least amount of positive sentiment) -- \textbf{the most negative term}:\\
1. th*nks \hspace{2mm} 2. doesn't work \hspace{2mm} 3. w00t \hspace{2mm} 4. \#theworst\\[-10pt]
\noindent\makebox[\linewidth]{\rule{0.49\textwidth}{0.4pt}}
}
\vspace*{1mm}
\noindent Each 4-tuple was annotated by ten respondents. 

The responses were then translated into real-valued scores and also a ranking of terms by sentiment for all the terms through a simple counting procedure: For
each term, its score is calculated as the percentage of times the term was chosen as the most positive minus the percentage of times the term was chosen as the most negative \cite{Orme_2009,flynn2014}.
The scores range from -1 (the most negative) to 1 (the most positive).

\subsection{Lexicons Created With Best--Worst Scaling}

\noindent {\scshape SemEval-2015 English Twitter Lexicon:} This lexicon is comprised of 1,515 high-frequency English single words and simple negated expressions commonly found in tweets. 
The set includes regular English words as well as some misspelled words (e.g., {\it parlament}), creatively-spelled words (e.g., {\it happeee}), hashtagged words (e.g., {\it \#loveumom}), and emoticons. 
\vspace{1mm}

\noindent {\scshape SemEval-2016 Arabic Twitter Lexicon:} This lexicon was created in a similar manner as the English Twitter Lexicon but using Arabic words and negated expressions commonly found in Arabic tweets.  
It has 1,367 terms.
\vspace{1mm}

\noindent {\scshape SemEval-2016 General English Sentiment Modifiers Lexicon} aka {\scshape Sentiment Composition Lexicon for Negators, Modals, and Degree Adverbs (SCL-NMA):} This lexicon consists of all 1,621 positive and negative single words from Osgood's seminal study on word meaning \cite{Osgood57} available in General Inquirer \cite{Stone66}. 
In addition, it includes 1,586 high-frequency phrases formed by the Osgood words in combination with simple negators such as {\it no}, {\it don't}, and {\it never}, modals such as {\it can}, {\it might}, and {\it should}, or degree adverbs such as {\it very} and {\it fairly}.  
More details on the lexicon creation and an analysis of the effect of different modifiers on sentiment can be found in \cite{SCL-NMA}.

\vspace{1mm}
\noindent Table~\ref{tab:lex-examples} shows example entries from each lexicon.
The complete lists of modifiers used in the three lexicons are available 
online.\footnote{
www.saifmohammad.com/WebPages/SCL.html\#ETSL\\
\indent \indent \hspace{-1.3mm} www.saifmohammad.com/WebPages/SCL.html\#ATSL\\
\indent \indent \hspace{-1.3mm} www.saifmohammad.com/WebPages/SCL.html\#NMA}
Details on the use of these lexicons in SemEval shared tasks can be found in \cite{Rosenthal-EtAl:2015:SemEval,SemEval2016Task7}.

\setlength{\tabcolsep}{1pt}

\setcode{utf8}
\renewcommand{\arraystretch}{0.95}

\begin{table}[t!]
\caption{\label{tab:lex-examples} {Example entries from the three lexicons.}
}
\begin{center}
\resizebox{0.4\textwidth}{!}{
\begin{tabular}{lr}
\hline {\bf Lexicon, Term} & {\bf Sentiment}\\
 \hline
{\it SemEval-2015 English Twitter Lexicon}\\
\,\,\,\,\, yummm & 0.813\\
\,\,\,\,\, cant waitttt & 0.656\\
\,\,\,\,\, \#feelingsorryformyself & -0.547\\
\,\,\,\,\, :'( & -0.563\\[5pt]
{\it SemEval-2016 Arabic Twitter Lexicon}\\
\,\,\,\,\, \<الزوجية>\_\<السعادة>\# (marital happiness) & 0.800\\
\,\,\,\,\, \<يقين>\# (certainty) & 0.675\\
\,\,\,\,\, \<لا امكن> (not possible) & -0.400\\
\,\,\,\,\, \<ارهاب> (terrorism) & -0.925\\[5pt]
{\it SemEval-2016 General English Lexicon}\\
\,\,\,\,\, would be very easy & 0.431\\
\,\,\,\,\, did not harm & 0.194\\
\,\,\,\,\, increasingly difficult & -0.583\\
\,\,\,\,\, severe & -0.833\\
\hline
\end{tabular}
}
\end{center}
\vspace*{-4mm}
\end{table}

\setlength{\tabcolsep}{6pt}

\section{Quality of Annotations}
\label{analysis-maxdiff}

\subsection{Agreement and Reproducibility}

Let {\it majority answer} refer to the option chosen most often for a BWS question.
The percentage of responses that matched the majority answer were as follows: 82\% for the English Twitter Lexicon, 80\% for the Arabic Twitter Lexicon, and 80\% for the General English Lexicon.

Annotations are reliable if similar results are obtained from repeated trials. 
To test the reliability of our annotations, 
we randomly divided the sets of ten responses to each question into two halves and compared the rankings obtained from these two groups of responses.
The Spearman rank correlation coefficient between the two sets of rankings produced for each of the three lexicons was found to be at least 0.98. 
(The Pearson correlation coefficient between the two sets of sentiment scores for each lexicon was also at least 0.98.)
Thus, even though annotators might disagree about answers to individual questions, the aggregated scores produced by applying the counting procedure 
on the BWS annotations are remarkably
reliable at ranking terms. 

\textbf{Number of annotations needed:} 
Even though we obtained ten annotations per BWS question, we wanted to determine the least number of annotations needed
to obtain reliable sentiment scores. For every $k$ (where $k$ ranges from 1 to 10), we made the following calculations:
for each BWS question, we randomly selected $k$ annotations and calculated sentiment scores based on the selected subset of annotations. 
We will refer to these sets of scores for the different values of $k$ as $S_1$, $S_2$, and so on until $S_{10}$.
This process was repeated ten times for each $k$. 
The average Spearman rank correlation coefficient between $S_1$ and $S_{10}$ was 0.96, between $S_2$ and $S_{10}$ was 0.98, and $S_3$ and $S_{10}$ was 0.99.
This shows that as few as two or three annotations per BWS question are sufficient to obtain reliable sentiment scores.
Note that with $2\times n$ BWS questions (for $n$ terms), each term occurs in eight 4-tuples on average, and so even just one annotation per BWS question means that each term is assessed eight times.

\begin{figure}[t]
\centering
\includegraphics[width=2.3in]{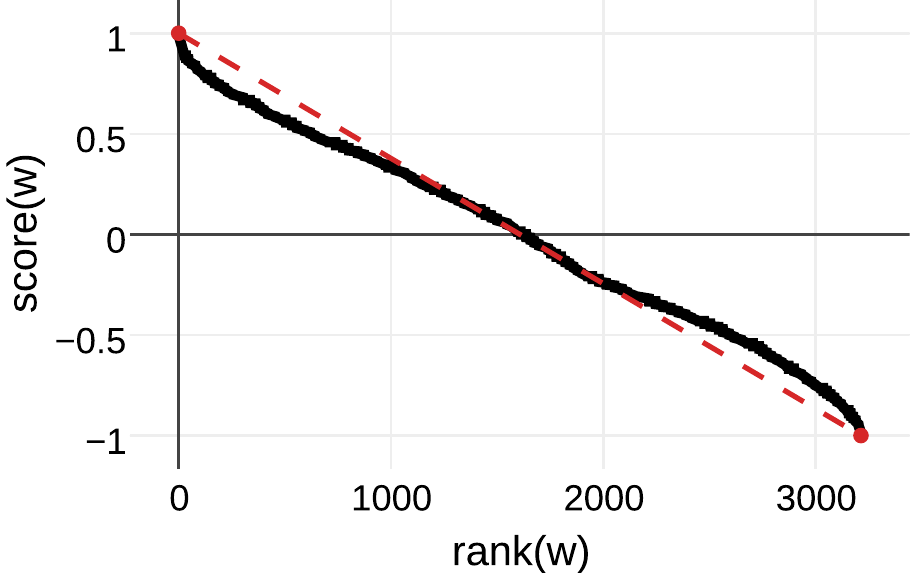}
\caption{\small Rank vs.\@ sentiment scores in SCL-NMA.}
\label{fig:OLex}
\vspace*{-3mm}
\end{figure}

\subsection{Distribution of Sentiment Scores}
Figure~\ref{fig:OLex} gives an overview of the sentiment scores in SCL-NMA. 
Each term in the lexicon is shown as a dot in the corresponding plot.
The x-axis is the rank of each term in the lexicon when the terms are ordered from most positive to least positive.
The y-axis is the real-valued sentiment score obtained from the BWS annotations.
Observe that the lexicon has entries for the full range of sentiment scores (-1 to 1); that is,
there are no significant gaps---ranges of sentiment scores for which the lexicon
does not include any terms.
The dashed red line indicates a uniform spread of scores, i.e., the same number of terms are expected to fall into each same-size interval of scores.
Observe that the lexicon has slightly fewer terms in the intervals with very high and very low sentiment scores.
Similar figures (not shown here) were obtained for the other two lexicons.

\subsection{Perception of Sentiment Difference}

The created lexicons capture sentiment associations at a fine level of granularity.
Thus, these annotations can help answer key questions such as: 
(1) If native speakers of a language are given two terms and asked which is more positive, 
how does human agreement vary with respect to the amount of difference in sentiment between the two focus terms?
It is expected that the greater the difference in sentiment, the higher the agreement, but the exact shape of this increase in agreement has not been shown till now.
(2) What least amount of difference in sentiment is perceptible to native speakers of a language?

\begin{figure}[t]
\centering
\includegraphics[width=2.5in]{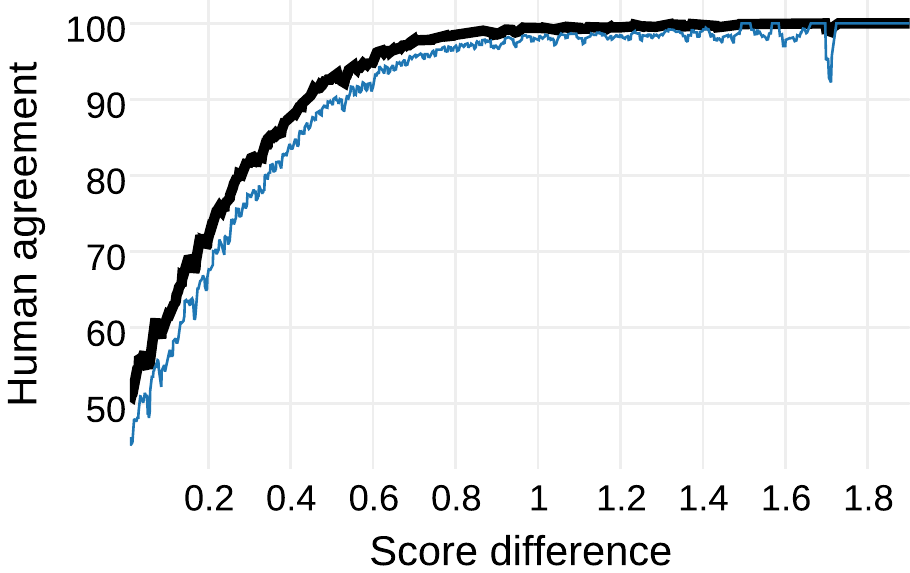}
\caption{\small SCL-NMA: Human agreement on annotating term $w_{1}$ as more positive than term $w_{2}$ for pairs with difference in scores $d$ = \score{w_{1}} - \score{w_{2}}.  
The x-axis represents $d$.
The y-axis plots the avg.\@ percentage of human annotations that judge term $w_{1}$ as more positive than term $w_{2}$ (thick line) and the corresponding 99.9\%-confidence lower bound (thin blue line).
}
\label{human-agreement}
\vspace*{-5mm}
\end{figure}

{\bf Agreement vs.\@ Sentiment Difference:} 
For each word pair $w_1$ and $w_2$ such that \score{w_{1}} $-$ \score{w_{2}} $\geq$ 0, 
we count the
number of BWS annotations from which we can infer that $w_{1}$ is more positive than $w_{2}$  
and divide this number by the total number of BWS annotations from which we can infer either that $w_{1}$ is more positive than $w_{2}$ or that $w_{2}$ is more positive than $w_{1}$.
(We can infer that $w_{1}$ is more positive than $w_{2}$ if in a 4-tuple that has both $w_1$ and $w_2$ the annotator
selected $w_{1}$ as the most positive or $w_{2}$ as the least positive. The case for $w_{2}$ being more positive than $w_{1}$ is similar.)
This ratio is the human agreement for $w_1$ being more positive than $w_2$, and we expect that it is correlated with the sentiment difference $d =$ \score{w_{1}} $-$ \score{w_{2}}.
To get more reliable estimates, we average the human agreement for all pairs of terms whose sentiment differs by $d \pm 0.01$. 
Figure~\ref{human-agreement} shows the resulting average human agreement on SCL-NMA.
Similar figures (not shown here) were obtained for the English and Arabic Twitter data.
Observe that the agreement grows rapidly with the increase in score differences.
Given two terms with sentiment differences of 0.4 or higher, more than 90\% of the annotators 
correctly identify the more positive term.

{\bf Least Difference in Sentiment that is Perceptible to Native Speakers:}
In psychophysics, there is a notion of {\it least perceptible difference} (aka {\it just-noticeable difference})---the amount by which something that can be measured (e.g., weight or sound intensity) needs to be changed in order for the difference to be noticeable by a human \cite{fechner1966elements}.
Analogously, we can measure the least perceptible difference in sentiment. 
If two words have close to identical sentiment associations, then
it is expected that native speakers will choose each of the words about the same number of times when forced to pick
a word that is more positive. However, as the difference in sentiment starts getting larger,
the frequency with which the two terms are chosen as most positive begins to diverge.
At one point, the frequencies diverge so much that we can say with high confidence that the two terms do not have the same sentiment associations.  
The average of this minimum difference in sentiment score is the least perceptible difference for sentiment. 
To determine the least perceptible difference, we first obtain the 99.9\%-confidence lower bounds on the human agreement (see the thin blue line in Figure~\ref{human-agreement}).  
The least perceptible difference is the point starting at which the lower bound consistently exceeds 50\% threshold (i.e., the point starting at which we observe with 99.9\% confidence that the human agreement is higher than chance).
The least perceptible difference when calculated from SCL-NMA is 0.069, from the English Twitter Lexicon is 0.080, and from the Arabic Twitter Lexicon is 0.087. 
Observe, that the estimates are very close to each other despite
being calculated from three completely independent datasets.
\newcite{SCL-NMA} use the least perceptible difference to determine whether a modifier significantly impacts the sentiment of the word it composes with.

\vspace*{-1mm}
\section{Conclusions}
\vspace*{-2mm}
We obtained real-valued sentiment association scores for single words and multi-word phrases in three domains (general English, English Twitter, and Arabic Twitter) by manual annotation and Best--Worst Scaling.   
Best--Worst Scaling exploits the comparative approach to annotation while keeping the number of annotations small. 
Notably, we showed that the procedure when repeated produces remarkably consistent rankings of terms by sentiment. 
This reliability allowed us to determine the value of the psycho-linguistic concept---least perceptible difference in sentiment.
We hope these findings will encourage further use of Best--Worst Scaling in linguistic annotation.

\bibliography{maxdiff}

\begin{thebibliography}{}

\bibitem[\protect\citename{Bradley and Lang}1999]{bradley1999affective}
Margaret~M Bradley and Peter~J Lang.
\newblock 1999.
\newblock Affective norms for {E}nglish words ({ANEW}): Instruction manual and
  affective ratings.
\newblock Technical report, The Center for Research in Psychophysiology,
  University of Florida.

\bibitem[\protect\citename{Cohen}2003]{Cohen_2003}
Steven~H. Cohen.
\newblock 2003.
\newblock Maximum difference scaling: Improved measures of importance and
  preference for segmentation.
\newblock Sawtooth Software, Inc.

\bibitem[\protect\citename{Esuli and Sebastiani}2006]{Esuli06}
Andrea Esuli and Fabrizio Sebastiani.
\newblock 2006.
\newblock {SENTIWORDNET}: A publicly available lexical resource for opinion
  mining.
\newblock In {\em Proceedings of the 5th Conference on Language Resources and
  Evaluation (LREC)}, pages 417--422.

\bibitem[\protect\citename{Fechner}1966]{fechner1966elements}
Gustav Fechner.
\newblock 1966.
\newblock {\em Elements of psychophysics. Vol. I.}
\newblock New York: Holt, Rinehart and Winston.

\bibitem[\protect\citename{Flynn and Marley}2014]{flynn2014}
T.~N. Flynn and A.~A.~J. Marley.
\newblock 2014.
\newblock Best-worst scaling: theory and methods.
\newblock In Stephane Hess and Andrew Daly, editors, {\em Handbook of Choice
  Modelling}, pages 178--201. Edward Elgar Publishing.

\bibitem[\protect\citename{Graham \bgroup et al.\egroup }2015]{grahamaccurate}
Yvette Graham, Nitika Mathur, and Timothy Baldwin.
\newblock 2015.
\newblock Accurate evaluation of segment-level machine translation metrics.
\newblock In {\em Proceedings of the Annual Conference of the North American
  Chapter of the ACL (NAACL)}, pages 1183--1191.

\bibitem[\protect\citename{Grijalva \bgroup et al.\egroup
  }2015]{grijalva2014gender}
Emily Grijalva, Daniel~A. Newman, Louis Tay, M.~Brent Donnellan, P.D. Harms,
  Richard~W. Robins, and Taiyi Yan.
\newblock 2015.
\newblock Gender differences in narcissism: A meta-analytic review.
\newblock {\em Psychological bulletin}, 141(2):261--310.

\bibitem[\protect\citename{Hartner}2013]{hartner2013lingering}
Marcus Hartner.
\newblock 2013.
\newblock The lingering after-effects in the reader's mind -- an investigation
  into the affective dimension of literary reading.
\newblock {\em Journal of Literary Theory Online}.

\bibitem[\protect\citename{Hu and Liu}2004]{Hu04}
Minqing Hu and Bing Liu.
\newblock 2004.
\newblock Mining and summarizing customer reviews.
\newblock In {\em Proceedings of the 10th ACM SIGKDD International Conference
  on Knowledge Discovery and Data Mining (KDD)}, pages 168--177, New York, NY,
  USA.

\bibitem[\protect\citename{Kiritchenko and Mohammad}2016]{SCL-NMA}
Svetlana Kiritchenko and Saif~M. Mohammad.
\newblock 2016.
\newblock The effect of negators, modals, and degree adverbs on sentiment
  composition.
\newblock In {\em Proceedings of the Workshop on Computational Approaches to
  Subjectivity, Sentiment and Social Media Analysis (WASSA)}.

\bibitem[\protect\citename{Kiritchenko \bgroup et al.\egroup
  }2014]{Kiritchenko2014}
Svetlana Kiritchenko, Xiaodan Zhu, and Saif~M. Mohammad.
\newblock 2014.
\newblock Sentiment analysis of short informal texts.
\newblock {\em Journal of Artificial Intelligence Research}, 50:723--762.

\bibitem[\protect\citename{Kiritchenko \bgroup et al.\egroup
  }2016]{SemEval2016Task7}
Svetlana Kiritchenko, Saif~M. Mohammad, and Mohammad Salameh.
\newblock 2016.
\newblock {SemEval-2016 Task 7}: Determining sentiment intensity of {E}nglish
  and {A}rabic phrases.
\newblock In {\em Proceedings of the International Workshop on Semantic
  Evaluation (SemEval)}, San Diego, California, June.

\bibitem[\protect\citename{Kleres}2011]{kleres2011emotions}
Jochen Kleres.
\newblock 2011.
\newblock Emotions and narrative analysis: A methodological approach.
\newblock {\em Journal for the Theory of Social Behaviour}, 41(2):182--202.

\bibitem[\protect\citename{Louviere and Woodworth}1990]{Louviere_1990}
Jordan~J. Louviere and George~G. Woodworth.
\newblock 1990.
\newblock Best-worst analysis.
\newblock Working Paper.
\newblock Department of Marketing and Economic Analysis, University of Alberta.

\bibitem[\protect\citename{Louviere \bgroup et al.\egroup }2015]{Louviere2015}
Jordan~J. Louviere, Terry~N. Flynn, and A.~A.~J. Marley.
\newblock 2015.
\newblock {\em {Best-Worst Scaling}: Theory, Methods and Applications}.
\newblock Cambridge University Press.

\bibitem[\protect\citename{Mohammad and Kiritchenko}2015]{COIN:COIN12024}
Saif~M. Mohammad and Svetlana Kiritchenko.
\newblock 2015.
\newblock Using hashtags to capture fine emotion categories from tweets.
\newblock {\em Computational Intelligence}, 31(2):301--326.

\bibitem[\protect\citename{Mohammad and Turney}2013]{MohammadT13}
Saif~M. Mohammad and Peter~D. Turney.
\newblock 2013.
\newblock Crowdsourcing a word--emotion association lexicon.
\newblock {\em Computational Intelligence}, 29(3):436--465.

\bibitem[\protect\citename{Mohammad \bgroup et al.\egroup }2016a]{stance-lrec}
Saif~M. Mohammad, Svetlana Kiritchenko, Parinaz Sobhani, Xiaodan Zhu, and Colin
  Cherry.
\newblock 2016a.
\newblock A dataset for detecting stance in tweets.
\newblock In {\em Proceedings of 10th edition of the the Language Resources and
  Evaluation Conference (LREC)}, Portoro\v{z}, Slovenia.

\bibitem[\protect\citename{Mohammad \bgroup et al.\egroup }2016b]{MohammadSK16}
Saif~M. Mohammad, Parinaz Sobhani, and Svetlana Kiritchenko.
\newblock 2016b.
\newblock Stance and sentiment in tweets.
\newblock {\em Special Section of the ACM Transactions on Internet Technology
  on Argumentation in Social Media}, Submitted.

\bibitem[\protect\citename{Mohammad}2012]{Mohammad2012}
Saif~M Mohammad.
\newblock 2012.
\newblock From once upon a time to happily ever after: Tracking emotions in
  mail and books.
\newblock {\em Decision Support Systems}, 53(4):730--741.

\bibitem[\protect\citename{Orme}2009]{Orme_2009}
Bryan Orme.
\newblock 2009.
\newblock Maxdiff analysis: Simple counting, individual-level logit, and {HB}.
\newblock Sawtooth Software, Inc.

\bibitem[\protect\citename{Osgood \bgroup et al.\egroup }1957]{Osgood57}
Charles~E Osgood, George~J Suci, and Percy Tannenbaum.
\newblock 1957.
\newblock {\em The measurement of meaning}.
\newblock University of Illinois Press.

\bibitem[\protect\citename{Pontiki \bgroup et al.\egroup
  }2014]{Semeval2014task4}
Maria Pontiki, Harris Papageorgiou, Dimitrios Galanis, Ion Androutsopoulos,
  John Pavlopoulos, and Suresh Manandhar.
\newblock 2014.
\newblock {SemEval}-2014 {T}ask 4: Aspect based sentiment analysis.
\newblock In {\em Proceedings of the 8th International Workshop on Semantic
  Evaluation (SemEval)}, Dublin, Ireland.

\bibitem[\protect\citename{Rosenthal \bgroup et al.\egroup
  }2014]{rosenthal-EtAl:2014:SemEval}
Sara Rosenthal, Alan Ritter, Preslav Nakov, and Veselin Stoyanov.
\newblock 2014.
\newblock {SemEval-2014 Task 9}: Sentiment analysis in {Twitter}.
\newblock In {\em Proceedings of the 8th International Workshop on Semantic
  Evaluation (SemEval)}, pages 73--80, Dublin, Ireland, August.

\bibitem[\protect\citename{Rosenthal \bgroup et al.\egroup
  }2015]{Rosenthal-EtAl:2015:SemEval}
Sara Rosenthal, Preslav Nakov, Svetlana Kiritchenko, Saif Mohammad, Alan
  Ritter, and Veselin Stoyanov.
\newblock 2015.
\newblock {SemEval-2015 Task 10}: Sentiment analysis in {T}witter.
\newblock In {\em Proceedings of the 9th International Workshop on Semantic
  Evaluation (SemEval)}, pages 450--462, Denver, Colorado.

\bibitem[\protect\citename{Stone \bgroup et al.\egroup }1966]{Stone66}
Philip Stone, Dexter~C. Dunphy, Marshall~S. Smith, Daniel~M. Ogilvie, and
  associates.
\newblock 1966.
\newblock {\em The General Inquirer: A Computer Approach to Content Analysis}.
\newblock The MIT Press.

\bibitem[\protect\citename{Tang \bgroup et al.\egroup }2014]{tang2014building}
Duyu Tang, Furu Wei, Bing Qin, Ming Zhou, and Ting Liu.
\newblock 2014.
\newblock Building large-scale {T}witter-specific sentiment lexicon: A
  representation learning approach.
\newblock In {\em Proceedings of the International Conference on Computational
  Linguistics (COLING)}, pages 172--182.

\bibitem[\protect\citename{Warriner \bgroup et al.\egroup
  }2013]{warriner2013norms}
Amy~Beth Warriner, Victor Kuperman, and Marc Brysbaert.
\newblock 2013.
\newblock Norms of valence, arousal, and dominance for 13,915 {E}nglish lemmas.
\newblock {\em Behavior Research Methods}, 45(4):1191--1207.

\bibitem[\protect\citename{Wilson \bgroup et al.\egroup }2005]{Wilson05}
Theresa Wilson, Janyce Wiebe, and Paul Hoffmann.
\newblock 2005.
\newblock Recognizing contextual polarity in phrase-level sentiment analysis.
\newblock In {\em Proceedings of the Joint Conference on HLT and EMNLP}, pages
  347--354, Stroudsburg, PA, USA.

\end{thebibliography}
\bibliographystyle{naaclhlt2016}

\end{document}